\documentclass[conference]{IEEEtran}
\IEEEoverridecommandlockouts
\usepackage{cite}
\usepackage{amsmath,amssymb,amsfonts}
\usepackage{algorithmic}
\usepackage{graphicx}
\usepackage{textcomp}
\usepackage{xcolor}
\def\BibTeX{{\rm B\kern-.05em{\sc i\kern-.025em b}\kern-.08em
    T\kern-.1667em\lower.7ex\hbox{E}\kern-.125emX}}
\begin{document}

\title{Autonomous Underwater Cognitive System for Adaptive Navigation: A SLAM-Integrated Cognitive Architecture}

\author{
\IEEEauthorblockN{K. A. I. N Jayarathne}
\IEEEauthorblockA{\textit{Department of Computational} \\ \textit{Mathematics} \\ \textit{University of Moratuwa}\\ Moratuwa, Sri Lanka \\ jayarathnekain.22@uom.lk}
\and
\IEEEauthorblockN{R. M. N. M. Rathnayaka}
\IEEEauthorblockA{\textit{Department of Computational} \\ \textit{Mathematics} \\ \textit{University of Moratuwa}\\ Moratuwa, Sri Lanka \\ rathnayakarmnm.22@uom.lk}
\and
\IEEEauthorblockN{D. P. S. S. Peiris}
\IEEEauthorblockA{\textit{Department of Computational} \\ \textit{Mathematics} \\ \textit{University of Moratuwa}\\ Moratuwa, Sri Lanka \\ peirisdpss.22@uom.lk}
}

\maketitle

\begin{abstract}
Deep-sea exploration faces critical challenges including disorientation, communication loss, and navigational failures in hostile underwater environments. This paper presents an Autonomous Underwater Cognitive System (AUCS) that integrates Simultaneous Localization and Mapping (SLAM) with a Soar-based cognitive architecture to enable adaptive navigation under dynamic oceanic conditions. The system combines multi-sensor fusion (SONAR, LiDAR, IMU, DVL) with cognitive reasoning capabilities including perception, attention, planning, and learning. Unlike conventional reactive SLAM systems, AUCS incorporates semantic understanding, adaptive sensor management, and memory-based learning to distinguish between dynamic and static objects, thus reducing false loop closures and improving long-term map consistency. The proposed architecture demonstrates a complete perception-cognition-action-learning cycle, enabling autonomous underwater vehicles to sense, reason, and adapt intelligently in challenging marine environments. This work addresses critical safety limitations observed in previous deep-sea missions and establishes a foundation for next-generation cognitive submersible systems.
\end{abstract}


\section{Introduction}
The exploration of the deep ocean remains one of the most complex and challenging frontiers in robotics and marine technology. Despite significant advances, submersible operations continue to face challenges such as disorientation, communication loss, and catastrophic mechanical or navigational failures. These challenges are amplified by the hostile underwater environment, where extreme pressure, limited visibility, and dynamic topographies demand precise, adaptive, and intelligent control systems.

The tragic loss of the OceanGate Titan submarine in 2023 highlighted the urgent need for robust safety mechanisms and intelligent situational awareness in deep-sea missions \cite{oceangate2023}. Traditional underwater navigation systems are mainly based on static control algorithms and isolated sensor inputs, which limit their adaptability to unpredictable conditions. Moreover, most current submarines lack cognitive reasoning capabilities that can integrate environmental, structural, and operational data for real-time decision making.

This limitation underscores the need for an autonomous cognitive submarine system that not only navigates independently but also learns, perceives, and reacts dynamically to changing underwater conditions. Autonomous Underwater Vehicles (AUVs) have been extensively studied for underwater exploration \cite{paull2014}, but typically lack the cognitive reasoning layer necessary for truly adaptive behavior in complex scenarios.

This research aims to design and propose a human-in-the-loop autonomous cognitive submarine system that integrates advanced navigation sensors, Artificial Intelligence (AI), and a pretrained naval knowledge base to ensure precise underwater navigation and minimize disorientation risks. The system improves route planning, obstacle avoidance, and real-time decision-making during deep-sea exploration.

By combining sensor fusion technologies such as LIDAR, SONAR, pressure sensors, Doppler Velocity Log (DVL), and Inertial Measurement Unit (IMU) sensors with the SLAM mechanism and an AI-driven decision-making framework, the proposed system prioritizes safety, reliability, and operational resilience. The integration of cognitive features such as perception, attention and focus, reasoning and decision making, planning, autonomy, adaptation and learning distinguishes this system from conventional approaches.

Ultimately, this study seeks to address the technological and safety limitations observed in prior missions, offering a foundation for next-generation submersible systems capable of autonomous, intelligent, and secure deep-sea navigation.

\section{Literature Review}

\subsection{SLAM Technologies for Underwater Environments}
Simultaneous Location Mapping (SLAM) has been widely studied for autonomous navigation in challenging environments, including underwater scenarios where GPS signals are unavailable and visibility conditions are highly variable. Traditional SLAM methods, originally developed for terrestrial or aerial environments, face unique constraints in underwater applications such as light attenuation, image distortion, turbidity, and the presence of dynamic marine elements \cite{wang2023slam}.

Recent research has increasingly focused on multi-sensor fusion approaches and domain-adaptive optimization techniques to enhance robustness and accuracy. Wang et al. \cite{wang2023slam} present a comprehensive overview of key SLAM technologies adapted for underwater environments, illustrating the SLAM pipeline from sensor acquisition to back-end optimization and loop closure. Their survey emphasizes the importance of multi-sensor fusion, particularly the integration of visual, sonar, inertial, and depth data to overcome the limitations of any single sensing modality.

The authors categorize fusion techniques into loose, tight, and ultra-tight coupling strategies and highlight the necessity of efficient data association and preprocessing, such as image enhancement and color correction, to maintain localization accuracy. Furthermore, they identify critical challenges that remain unsolved in the field, namely the lack of semantic understanding in mapping, the absence of adaptive decision-making during exploration, and the limited capability for real-time computation on resource-constrained underwater vehicles.

\subsection{Multi-Sensor Fusion Approaches}
Rahman et al. \cite{rahman2019svin2} introduce SVIn2, a tightly-coupled stereo visual-inertial-sonar SLAM system specifically designed for underwater robotics. Their method combines stereo vision, IMU, sonar, and depth sensor data within a unified optimization framework. The system includes several innovations: a robust two-step scale refinement process that uses depth readings to correct metric drift, image preprocessing through Contrast Limited Adaptive Histogram Equalization (CLAHE) to counteract color degradation, and a Bag-of-Words (BoW) approach for loop closure and relocalization.

Experimental results demonstrate that SVIn2 significantly improves localization robustness compared to monocular or loosely-coupled SLAM systems under low-visibility and dynamically lit conditions. However, despite these improvements, the system remains reactive—it estimates and optimizes trajectories but lacks higher-level reasoning or learning capabilities to interpret or adapt to environmental semantics.

\subsection{Cognitive Architectures in Robotics}
Cognitive architectures provide a framework for integrating perception, reasoning, and action in autonomous systems. The Soar cognitive architecture \cite{laird2012soar} has been successfully applied to various robotic applications, offering mechanisms for goal-driven behavior, procedural learning through chunking, and working memory management. Recent work has explored the integration of cognitive architectures with robotic systems for enhanced autonomy \cite{anderson2004integrated}.

\subsection{Research Gap}
Both underwater SLAM studies converge on a common limitation: the absence of cognitive and semantic reasoning. Current systems rely primarily on geometric and photometric information without understanding the meaning of observed features. They cannot distinguish between dynamic and static objects, adapt sensor configurations based on environmental feedback, or reason about the long-term consistency of maps across missions. Moreover, the high computational cost of optimization and sensor fusion often restricts real-time deployment on small AUVs.

To address these limitations, the proposed research introduces an underwater cognitive SLAM system that integrates a reasoning layer on top of conventional multi-sensor fusion architectures. This cognitive layer performs semantic interpretation, adaptive sensor management, and intelligent loop closure validation.

\section{Methodology}

\subsection{System Architecture Overview}
The proposed Autonomous Underwater Cognitive System (AUCS) is designed to enable an underwater vehicle to perform adaptive navigation, environmental mapping, and decision-making in dynamic and uncertain marine environments. The system integrates a multi-layered architecture consisting of three core modules: the Sensing and Data Acquisition Subsystem, the Simultaneous Localization and Mapping (SLAM) Module, and the Cognitive Control Layer based on the Soar Cognitive Architecture.

\begin{figure}[htbp]
\centerline{\includegraphics[width=\columnwidth]{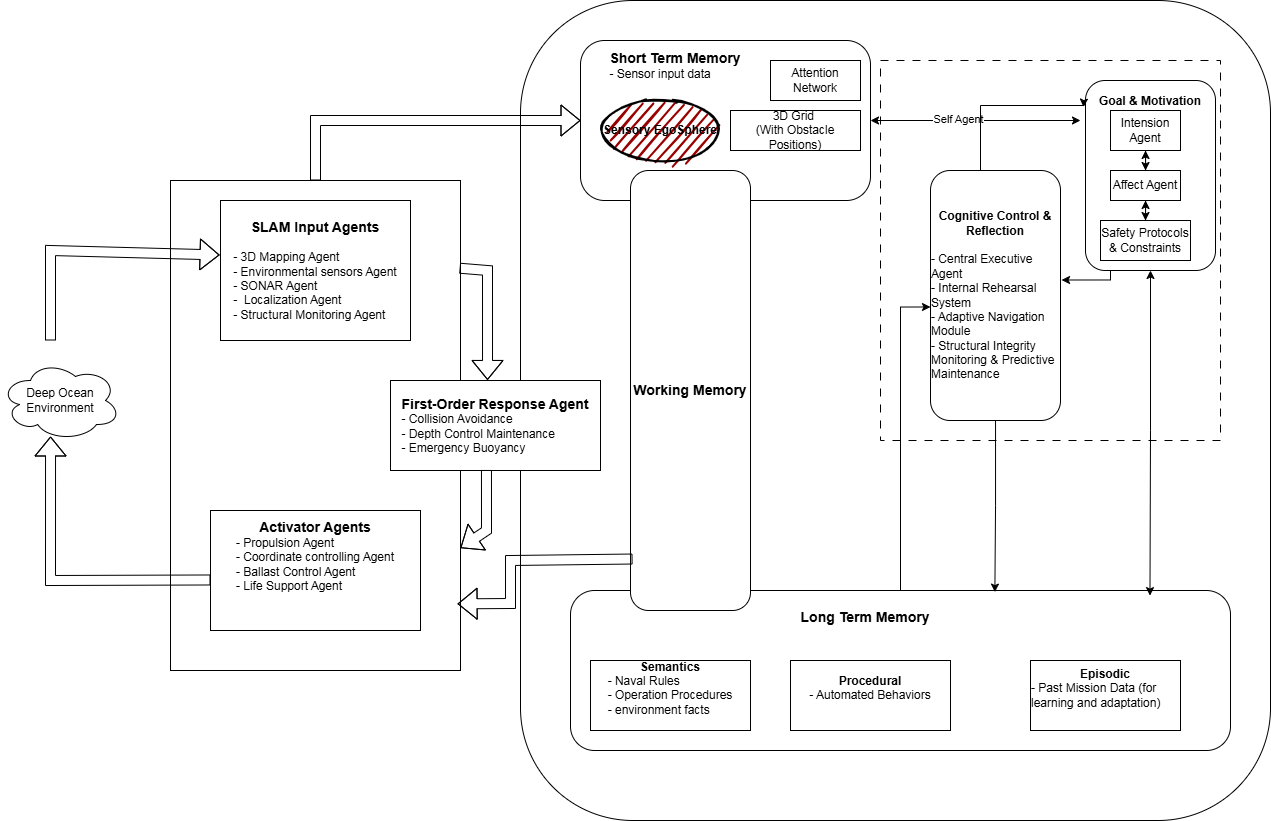}}
\caption{AUCS System Architecture showing integration of SLAM, memory systems, cognitive control, and actuator agents.}
\label{fig:architecture}
\end{figure}

The overall architecture (as shown in Fig.~\ref{fig:architecture}) consists of six major modules:
\begin{enumerate}
\item SLAM Input Agents
\item First-Order Response Agent
\item Activator Agents
\item Memory System (Short-Term, Working, Long-Term)
\item Cognitive Control and Reflection Unit
\item Goal and Motivation System
\end{enumerate}

Each module cooperatively works to achieve perception, reasoning, decision-making, and action execution in a closed-loop manner.

\subsection{SLAM Input Agents}
The SLAM Input Agents are responsible for gathering real-time data from the underwater environment using multiple sensor modalities:

\textbf{3D Mapping Agent:} Generates a three-dimensional point cloud representation of the environment using LiDAR and stereo vision systems.

\textbf{Environmental Sensor Agent:} Monitors temperature, pressure, salinity, and chemical properties of surrounding water to assess operational conditions.

\textbf{SONAR Agent:} Detects obstacles and measures distances using sound wave reflection, providing robust sensing in turbid conditions.

\textbf{Acoustic Sensor:} Supports SONAR data and provides fine-grained proximity detection for near-field obstacle avoidance.

\textbf{Structural Monitoring Agent:} Monitors the submarine's physical integrity and detects external impacts or structural stress.

All collected data are transmitted to Short-Term Memory for immediate processing and grid-based mapping.

\begin{figure}[htbp]
\centerline{\includegraphics[width=\columnwidth]{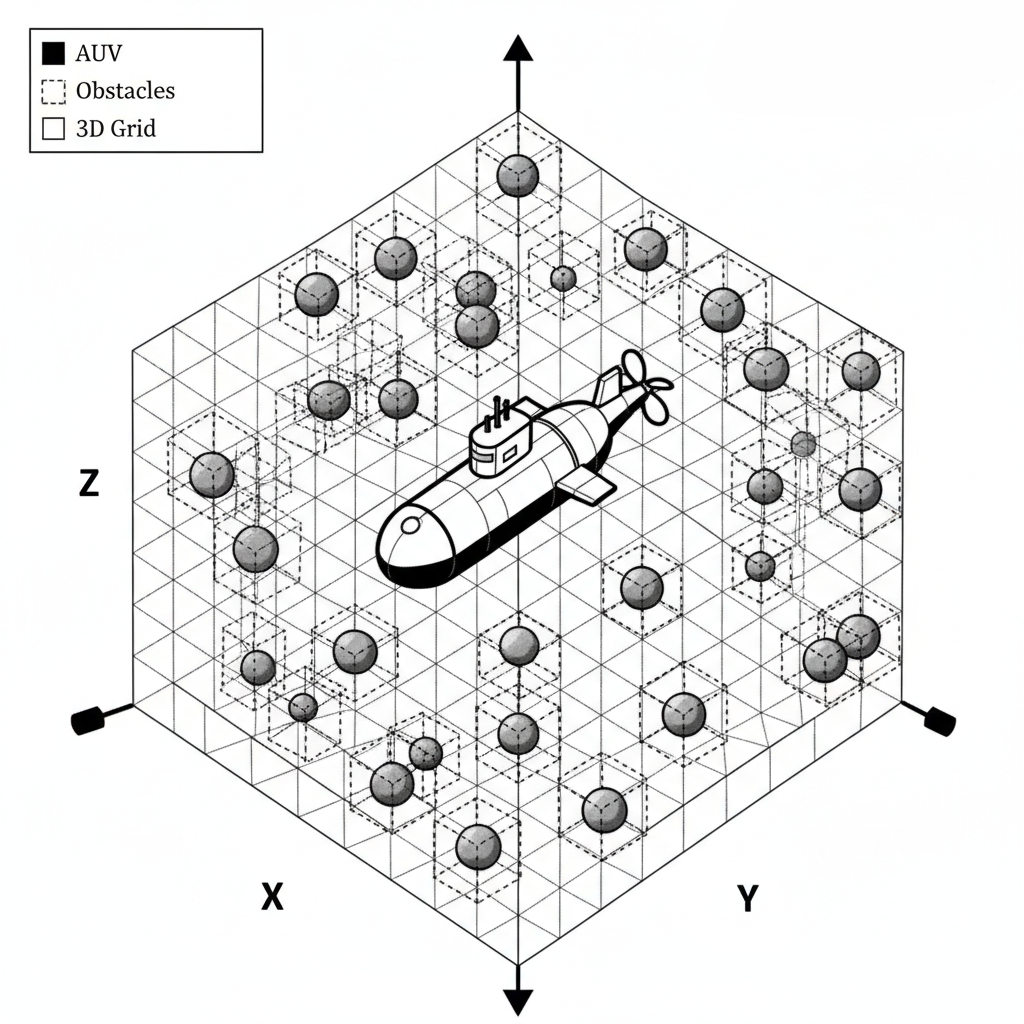}}
\caption{Submarine generating 3D grid mesh with detected obstacles marked in red, showing real-time environmental mapping and navigation space.}
\label{fig:3dgrid}
\end{figure}

As illustrated in Fig.~\ref{fig:3dgrid}, the submarine continuously generates and updates a 3D occupancy grid where obstacles detected by the sensor suite are marked and tracked in real-time.

\subsection{Memory System Architecture}

\subsubsection{Short-Term Memory (STM)}
STM temporarily stores the incoming sensor fusion data. The Sensory Edge Sphere module processes immediate sensory information, while the Attention Network filters important environmental features based on relevance and urgency.

A 3D Grid Map is dynamically generated and updated in real-time, storing detected obstacle positions and environmental features. This is where the SLAM algorithm operates, continuously performing two main tasks:

\textbf{Localization:} Estimating the submarine's position and orientation using sensor data and motion feedback from IMU and DVL.

\textbf{Mapping:} Building and refining a consistent 3D map of the environment based on sensor readings.

The algorithm applies a predict-update loop:
\begin{itemize}
\item \textit{Prediction step:} Uses the submarine's motion model to predict the next pose based on odometry.
\item \textit{Update step:} Incorporates new sensor data from SONAR, cameras, or acoustic sensors to correct the predicted pose and update the map using probabilistic estimation.
\end{itemize}

The resulting 3D grid with obstacles is then transferred to Working Memory for decision processing.

\subsubsection{Working Memory}
The Working Memory acts as an intermediate buffer connecting short-term perception and higher-level cognitive reasoning. It provides the First-Order Response Agent with situational awareness and enables immediate reactive behaviors such as:
\begin{itemize}
\item Collision Avoidance
\item Depth Control Maintenance
\item Emergency Buoyancy Adjustment
\end{itemize}

These reflexive responses ensure submarine safety in real time, based on SLAM data and pre-stored operational rules.

\subsubsection{Long-Term Memory (LTM)}
The LTM serves as a knowledge base that stores previously acquired experiences and learned operational rules:

\textbf{Semantic Memory:} Contains facts such as naval protocols, underwater navigation rules, and environmental characteristics.

\textbf{Procedural Memory:} Stores learned control policies and automated response sequences developed through experience.

\textbf{Episodic Memory:} Retains mission-specific data for learning and adaptation across deployments, enabling the system to recall and apply previous successful strategies.

During missions, relevant knowledge from LTM is retrieved into Working Memory to assist real-time reasoning and decision-making.

\subsection{Cognitive Control and Reflection}
This layer performs higher-order decision-making and adaptive reasoning, analogous to the human prefrontal cortex. Key components include:

\textbf{Central Executive Agent:} Coordinates all subsystems and allocates attention resources based on current mission priorities and environmental demands.

\textbf{Internal Rehearsal System:} Simulates possible future states before executing actions, allowing the system to evaluate potential outcomes and select optimal strategies.

\textbf{Adaptive Navigation Module:} Plans and adjusts navigation paths using SLAM's 3D grid and learned patterns from past missions, incorporating obstacle avoidance and energy optimization.

\textbf{Structural Integrity Monitoring:} Uses feedback from structural agents to ensure safe operation under pressure changes and external forces.

The cognitive control layer interacts with the Goal and Motivation modules to align actions with mission objectives while maintaining safety constraints.

\subsection{Goal and Motivation System}
The Goal and Motivation system manages high-level mission planning:

\textbf{Intention Agent:} Defines mission targets such as mapping a specific region, collecting samples, or avoiding hazardous zones.

\textbf{Affect Agent:} Evaluates current mission performance and updates emotional-like states representing risk level, urgency, and confidence.

\textbf{Safety Protocols and Constraints:} Ensure that all decisions meet operational safety limits, preventing risky maneuvers or actions that could lead to structural failures.

This subsystem can dynamically adjust goals based on environmental changes detected through SLAM, enabling adaptive mission execution.

\subsection{Activator Agents}
These agents perform the final stage of the perception-action loop:

\textbf{Propulsion Agent:} Executes motion commands and adjusts thrust levels.

\textbf{Coordinate Controlling Agent:} Manages direction and velocity adjustments for precise navigation.

\textbf{Ballast Control Agent:} Maintains depth and buoyancy through water ballast management.

They operate under directives from the Cognitive Control module, guided by the SLAM-based navigation map.

\subsection{Adaptive SLAM Feedback Loop}
The integration of SLAM into the cognitive framework enables an adaptive feedback loop:
\begin{enumerate}
\item SLAM generates an updated environmental model based on current sensor data.
\item Cognitive modules analyze the grid to assess navigability and detect possible hazards.
\item Updated control commands are issued to the Activator Agents for propulsion, ballast control, and maneuvering.
\item The resulting new position data are re-fed into SLAM to improve mapping accuracy.
\end{enumerate}

This continuous feedback ensures both situational awareness and intelligent adaptation in complex underwater terrains.

\subsection{Complete System Workflow}
AUCS begins operation by interacting with the underwater environment through a suite of heterogeneous sensors. All collected data is stored in Short-Term Memory for initial preprocessing. The Attention Network filters redundant information, prioritizing essential inputs such as nearby obstacles, which are fused into a composite 3D Grid Map.

The SLAM module simultaneously estimates the submarine's position using odometry from IMU and DVL, combined with visual and SONAR feedback. Probabilistic estimation mechanisms correct positional drift and update the map in real time.

Updated sensory information is transferred to Working Memory, where the Immediate Response Mechanism continuously monitors critical conditions to execute reflexive actions through Activator Agents. Cognitive Control processes this data to plan adaptive navigation, simulate future actions, and monitor structural integrity.

The Goal and Motivation Module aligns decision-making with mission objectives, evaluating risk and enforcing operational constraints. Once decisions are approved, Activator Agents execute precise motor commands. New sensory data is continuously fed back to STM, creating a closed perception-cognition-action-learning cycle.

\section{Discussion}

\subsection{Cognitive Reasoning Integration}
The proposed AUCS represents a significant advancement toward achieving autonomous and self-correcting submersible navigation in dynamic oceanic environments. Unlike conventional underwater SLAM or control systems, which are purely reactive, AUCS introduces a cognitive reasoning level that bridges low-level perception and high-level decision-making.

Among the standout contributions is the synergistic relationship between SLAM and cognition. Traditional SLAM pipelines, though effective for geometric localization, are not capable of interpreting the semantic and contextual meaning of underwater objects. The AUCS framework introduces semantic labeling and adaptive sensor management, enabling the system to distinguish between dynamic and static objects.

For example, by defining coral reefs as fixed obstacles and sea creatures as mobile objects, the system can avoid spurious loop closures and ensure enhanced long-term map consistency. This semantic reasoning directly addresses the limitations indicated by Wang et al. \cite{wang2023slam} and Rahman et al. \cite{rahman2019svin2}, who highlighted the lack of adaptive intelligence and environmental awareness in current SLAM systems.

\subsection{Multi-Level Cognitive Framework}
A key advantage of AUCS is its multi-level cognitive framework, drawing from human cognitive models such as the Soar architecture \cite{laird2012soar}. The separation of short-term, working, and long-term memory provides an organized system for real-time perception, decision-making, and learning.

Working Memory enables immediate responses such as collision avoidance and buoyancy control, while the Cognitive Control Layer facilitates higher-level reasoning, goal prioritization, and mission planning. Experiences are retained in Long-Term Memory over time, allowing the submarine to adjust its navigational strategies based on past operations—evidence of true cognitive adaptation.

\subsection{Operational Advantages}
From an operational perspective, the adaptive SLAM feedback loop ensures perception-action synchrony at all times. The cognitive level analyzes the grids that SLAM generates and dynamically updates propulsion, depth, and orientation commands. This closed-loop mechanism not only provides better situational awareness but also enables self-correcting behavior where the system learns to correct itself from environmental feedback.

This adaptability is necessary to cope with unpredictable underwater conditions such as turbidity variations, current changes, or unexpected obstacles. The system's ability to simulate future states through internal rehearsal before committing to actions significantly reduces the risk of hazardous maneuvers.

\subsection{Implementation Challenges}
However, several practical challenges must be addressed for real deployment. First, the computational expense of integrating multi-sensor fusion with cognitive reasoning challenges onboard processors, especially on small AUVs. Real-time operation will require efficient task scheduling, hardware acceleration using GPU or TPU processing units, or distributed edge-cloud cooperation.

Second, data reliability and sensor noise remain significant constraints. Acoustic backscattering, biofouling, or pressure-induced drift can compromise mapping accuracy. Robust sensor calibration and adaptive noise modeling are therefore essential for maintaining system performance.

Third, ensuring stability in cognitive decision loops when facing conflicting stimuli (such as simultaneous obstacle detection and goal seeking) will necessitate careful control-theoretic tuning between full autonomy and human-in-the-loop intervention.

\subsection{Broader Applications}
The AUCS architecture lays the foundation for next-generation cognitive marine robotics. Beyond exploration, these systems can have significant applications in subsea infrastructure inspection, search and rescue operations, environmental monitoring, and defense applications. Under human-in-the-loop monitoring, the model ensures that final mission-critical decisions such as emergency resurfacing or major course corrections still benefit from human judgment while maintaining autonomous efficiency.

Moreover, the introduction of memory-based learning and semantic reasoning paves the way for lifetime adaptive systems, where underwater vehicles progressively enhance their understanding of diverse marine terrain through successive missions.

\section{Conclusion}
This paper presented and validated the methodology for an Autonomous Underwater Cognitive System for Adaptive Navigation. The core contribution is the robust integration of a SLAM module with a high-level Soar-based Cognitive Architecture. The system demonstrates the ability to concurrently build a 3D Grid Map of unknown environments and accurately localize itself using a heterogeneous sensor suite including SONAR, IMU, and DVL.

Crucially, the cognitive layer provides advanced functionalities such as planning, reasoning and decision-making, goal management, and experiential learning through chunking mechanisms. This enables the AUV to exhibit truly adaptive behaviors like dynamic route adjustment and autonomous collision avoidance in the challenging deep-ocean environment.

By effectively perceiving its surroundings and adapting its operational rules based on experience, this architecture offers a significant step toward developing highly autonomous and intelligent underwater vehicles capable of complex, unsupervised missions. The integration of semantic understanding, adaptive sensor management, and memory-based learning distinguishes AUCS from conventional reactive systems.

Future work will focus on implementing and testing the architecture in real or simulated oceanic environments to evaluate its scalability, computational complexity, and long-term learning capabilities. Additional research will explore optimization of the cognitive decision-making algorithms and investigation of distributed processing architectures for resource-constrained platforms.

Ultimately, this work makes a foundational contribution toward realizing cognitive submersibles that can sense, reason, and act intelligently in the most challenging regions of the world's oceans, potentially transforming deep-sea exploration and underwater robotics.

\section*{Acknowledgment}
The authors would like to express their sincere appreciation to Dr.~Menaka Ranasinghe, Senior Lecturer at the Open University of Sri Lanka, for her invaluable guidance, constructive feedback, and continuous support throughout the development of this research. The authors also extend their gratitude to the Robotic Gen Team for providing valuable insights and technical knowledge related to SLAM and robot navigation, which significantly contributed to strengthening the practical foundation of this work. Furthermore, the authors are grateful to the anonymous reviewers for their insightful comments and suggestions, which helped improve the quality of this paper.

\end{document}